\documentclass{article}

\PassOptionsToPackage{numbers, compress}{natbib}

\usepackage[preprint]{neurips_2021}

\usepackage[utf8]{inputenc} 
\usepackage[T1]{fontenc}    
\usepackage{hyperref}       
\usepackage{url}            
\usepackage{booktabs}       
\usepackage{amsfonts}       
\usepackage{nicefrac}       
\usepackage{microtype}      
\usepackage{xcolor}         
\usepackage{graphicx}       
\usepackage{subcaption}     
\usepackage{mathtools}      
\usepackage{natbib}         
\usepackage{tablefootnote}  
\usepackage{longtable}      

\title{More layers! End-to-end regression and uncertainty on tabular data  with deep learning}

\author{%
  Ivan Bondarenko \textsuperscript{1, 2} \\
  \textsuperscript{1} Novosibirsk State University \\
  \textsuperscript{2} Huawei Novosibirsk Research Center \\
  Novosibirsk, Russia \\
  \texttt{i.bondarenko@g.nsu.ru} \\
}

\begin{document}

\maketitle

\begin{abstract}
  This paper attempts to analyze the effectiveness of deep learning for tabular data processing. It is believed that decision trees and their ensembles is the leading method in this domain, and deep neural networks must be content with computer vision and so on. But the deep neural network is a framework for building gradient-based hierarchical representations, and this key feature should be able to provide the best processing of generic structured (tabular) data, not just image matrices and audio spectrograms. This problem is considered through the prism of the Weather Prediction track in the Yandex Shifts challenge (in other words, the Yandex Shifts Weather task). This task is a variant of the classical tabular data regression problem. It is also connected with another important problem: generalization and uncertainty in machine learning. This paper proposes an end-to-end algorithm for solving the problem of regression with uncertainty on tabular data, which is based on the combination of four ideas: 1) deep ensemble of self-normalizing neural networks, 2) regression as parameter estimation of the Gaussian target error distribution, 3) hierarchical multitask learning, and 4) simple data preprocessing. Three modifications of the proposed algorithm form the top-3 leaderboard of the Yandex Shifts Weather challenge respectively. This paper considers that this success has occurred due to the fundamental properties of the deep learning algorithm, and tries to prove this.
\end{abstract}

\section{Introduction}

Deep learning is a very effective technique for such tasks, as computer vision, automatic speech recognition and natural language understanding. The "classic" ML solutions in these areas were based on pipelines of independent feature extractors, data preprocessors and dimensionality reductors, each of them usually based either on domain-specific heuristics or on locally optimal choice. A linear or "flat" nonlinear estimator, such as SVM, was used at the end of this pipeline. For example, the ILSVRC 2010 winner had the architecture of this type \cite{YangLLZYDCH10}. Contrary to this approach, any deep learning solution uses a very "pure" feature preprocessing. The key nuance of deep learning is a pipeline of hierarchical feature transformations (as hidden layers) with final classifier/regressor (as output layer), and all of them are trained together to minimize the joint loss function using the gradient descent. Since 2012, all ILSVRC winners are inspired by deep learning \cite{Krizhevsky2012}. We can see a similar trend in other ML domains, such as automatic speech recognition or natural language understanding. But not in a tabular data processing. Why? In the recent years some interesting papers about neural networks for tabular data processing were published \cite{Klambauer2017,Buturovic2020,Arik2021tabnet}, but gradient boosted decision trees and other techniques based on decision tree ensembles are still very widespread. Some researchers think that speech and vision data has a local ordering (neighbour pixels of a raster matrix or neighbour spectrums of a spectrogram are correlated), but tabular data does not have this property (the column order in a table does not matter). Hence, they postulate that neural networks which can detect such a local order using shared weights (by receptive fields in convolutional networks or by time in recurrent networks) have less effectiveness on any data without a local order compared to decision trees. However, there is reason to believe that it is not true, and the key advantage of deep learning is based on another thing, namely the above-mentioned building of "natural" hierarchy of feature space transformations with end-to-end gradient learning.

Yet another issue to analyse is uncertainty. The truth is, machine learning is an inductive approach to building an artificial intelligence system. Consequently, training data defines the behaviour of such system on any new data. But the data source can be changed, and the next portion of data will be sampled from a different statistical population relative to training set. So, any AI algorithm has to estimate its own competence. In critical domains, such as medicine or finance, it must give predictions only if it is sure. Also, the algorithm should alert the operator if some of the input data is not typical for the task being solved. Aside from that, solving the uncertainty and generalization problem in machine learning also improves the robustness of the algorithm to adversarial attacks. Therefore, ability to model uncertainty and reject "unknown" entities is a significant feature of any machine learning system in practice. In this context, the Yandex Shifts challenge, devoted to robustness and uncertainty under real-world distributional shift, generates interesting and indicative tasks, including weather prediction that accounts for different climatic zones and time periods.\footnote{\url{https://research.yandex.com/shifts/weather}}

So, developing a deep learning algorithm that can either solve a regression task or estimate data uncertainty and self-competence seems to be an actual and non-trivial problem. And depth can help us approximate unknown dependencies between inputs and targets in a more efficient way than ensembles of decision trees and other ML approaches, some nuances notwithstanding.

\section{Task, quality measure and baseline}
\label{baseline}

Data for the Weather Prediction task is described in details in \cite{Malinin2021shifts}. Nevertheless, some important peculiarities should be mentioned.

\begin{figure*}[hbt!]
 \centering
\begin{minipage}[h]{0.8\linewidth}
  \centering
  \centerline{\includegraphics[width=13cm]{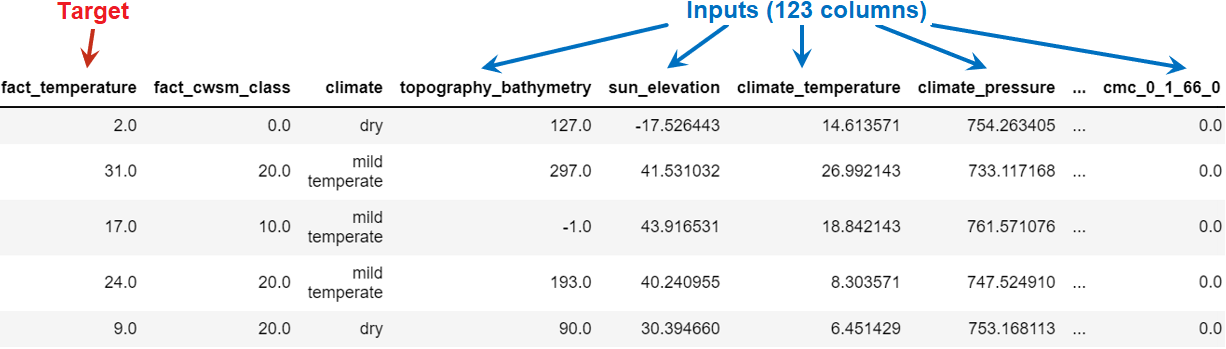}}
  \vspace{-0.1cm}
\end{minipage}
 \caption{Fragment of tabular data for the Weather Prediction task.}
 \label{fig:sample-data}
\end{figure*}

Data consist of 123 inputs, one target and some additional meta information (see Fig.~\ref{fig:sample-data}). Inputs include source measures by weather stations and multi-criterion forecasts by three systems:  Global Forecast System (GFS), the Canadian Meteorological Center (CMC), and the Weather Research and Forecasting (WRF). Target column contains the temperature that was actually registered, accurate within one degree Celsius. The meta information consists of several columns, including some time data, but time factor is not presented explicitly in the inputs. In that way, the Weather Prediction task is a regression on "typical" tabular data, without any time series.

All data samples are grouped by \textbf{inner} and \textbf{outer partitions}. The inner partition includes meteorological data from tropical, dry and mild temperature climate zones, and the outer partition consists of meteorological data collected at more inclement conditions: snow and polar. Also, the outer partition data differs in the time period of collection (the outer partition was collected later). Split into training, development and evaluation sub-sets is implemented with this grouping taken into account.

\begin{itemize}
  \item The training set includes the inner partition only, and is named as \verb+train+.
  \item The development set consists of data from both partitions (\verb+dev_in+ and \verb+dev_out+), and this set can be used for preliminary evaluation of proposed algorithm quality by challenge participants (but not for training directly or validation-in-training).
  \item The evaluation set includes the same data packages as the development set (names of the packages are \verb+eval_in+ and \verb+eval_out+), but it is larger and does not include targets; so, it is used for the final evaluation of the algorithm quality by the challenge organizers.
\end{itemize}

\begin{figure*}[hbt!]
 \centering
\begin{minipage}[h]{0.8\linewidth}
  \centering
  \centerline{\includegraphics[width=7.2cm]{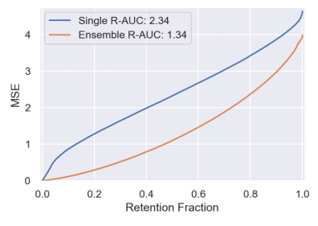}}
  \vspace{-0.1cm}
\end{minipage}
 \caption{Example MSE retention curves on evaluation dataset, reached by the baseline solution, proposed by organizers (for single CatBoost and ensemble of CatBoosts). \cite{Malinin2021shifts}}
 \label{fig:rauc-mse-example}
\end{figure*}

A joint assessment of uncertainty and robustness is based on \textbf{R-AUC MSE score} instead of the usual mean square error (MSE). This score is the main performance metric, and it is calculated as \textbf{the area under error-retention curve} (you can see the example on Fig.~\ref{fig:rauc-mse-example}). Error-retention curve is built as the MSE of all predictions after their retention by uncertainty value with different uncertainty thresholds (i.e., only predictions with uncertainties not exceeding the specified threshold are retained and are taken into consideration when calculating the MSE). The lower the value of the area under this curve, the better. R-AUC MSE was proposed as a metric for regression with uncertainty in \cite{Malinin2021shifts}, but evidently this assessment scenario was inspired by assessing misclassification detection via rejection curves in \cite{Malinin2019thesis}.

The baseline proposed by the organizers represents an ensemble of 10 gradient boosted decision trees, which are built on the basis of CatBoost algorithmic techniques \cite{Ostroumova2018catboost}. It is known that default hyper-parameters of CatBoost are determined very well, but the authors of the baseline proposed a set of special non-default values for increased effectiveness:

\begin{itemize}
  \item tree depth is 8 (default value is 6);
  \item 20000 trees in each CatBoost (an extremely large value, compared to the default value of 1000);
  \item a special loss function \verb+RMSEWithUncertainty+, proposed in \cite{Malinin2021gbtuncertainty} instead of the usual MSE or RMSE to consider the uncertainty in training and inference.
\end{itemize}

The above-mentioned loss function allows a trainable algorithm to estimate the best normal distribution described by our regression targets in the most likely way (instead of the “classic” average squared difference between targets and point-estimated outputs). It is formulated in the following way:

\begin{equation}
\label{eq:loss}
    L(\Theta|\mathcal{D}) = \mathbb{E}_{\mathcal{D}}\lbrack-\log\left(\mathrm{p}(y|x, \Theta)\right)\rbrack = -\frac{1}{N} \displaystyle\sum_{i=1}^{N} \log\left(\mathrm{p}(y^{(i)}|x^{(i)}, \Theta)\right),
\end{equation}

where:

\begin{equation}
\label{eq:normal-distr}
\mathrm{p}(y|x, \Theta) = \mathcal{N}(y|\mu, \sigma), \{\mu, \log(\sigma)\} = F(x).
\end{equation}

So, the CatBoost regressor \(F(x)\) predicts two values: the mean \(\mu\) and the logarithm of the standard deviation \(\sigma\) (prediction of the logarithm instead of the immediate value helps to provide only positive standard deviations).

\section{Proposed method and its quality}
\label{snn}

\subsection{Method}

A special deep ensemble (i.e. ensemble of deep neural networks) with uncertainty, hierarchical multitask learning and some other features is proposed. It is built on the basis of the following key techniques:

\begin{enumerate}
    \item A simple \textbf{preprocessing} is applied \textbf{to the input data}:
        \begin{itemize} 
            \item \emph{imputing}: the missing values are replaced in all input columns following a simple constant strategy (fill value is $-$1);
            \item \emph{quantization}: each input column is discretized into quantile bins, and the number of these bins is detected automatically; after that the bin identifier can be considered as a quantized numerical value of the original feature; 
            \item \emph{standardization}: each quantized column is standardized by removing the mean and scaling to unit variance;
            \item \emph{decorrelation}: all possible linear correlations are removed from the feature vectors discretized by above-mentioned way; the decorrelation is implemented using PCA \cite{esbensen2002multivariate}.
        \end{itemize} 
    \item An even simpler \textbf{preprocessing} is applied \textbf{to targets}: it is based only on removing the mean and scaling to unit variance.
    \item A \textbf{deep neural network} is built for regression with uncertainty:
        \begin{itemize} 
            \item \emph{self-normalization}: this is a self-normalized neural network, also known as SNN \cite{Klambauer2017};
            \item \emph{inductive bias}: neural network weights are tuned using a hierarchical multitask learning with the temperature prediction as a high-level regression task and the coarsened temperature class recognition as a low-level classification task;
            \item \emph{uncertainty}: a special loss function similar to \verb+RMSEWithUncertainty+ \cite{Malinin2021gbtuncertainty} is applied to training according to formulas (\ref{eq:loss}) and (\ref{eq:normal-distr}), but \(\sigma = \mathrm{softplus}(\tilde{\sigma})\) is estimated instead of \(\log(\sigma)\);
            \item \emph{robustness}: a supervised contrastive learning based on N-pairs loss \cite{Sohn2016ImprovedDM} is applied instead of the crossentropy-based classification as a low-level task in the hierarchical multitask learning; it provides more robustness of the trainable neural network \cite{Khosla2020npairs}.
        \end{itemize}
    \item A special technique of \textbf{deep ensemble} creation is implemented: it uses a model average approach like bagging \cite{Breiman2004BaggingP}, but new training and validation sub-sets for corresponding ensemble items are generated using stratification based on coarsened temperature classes.
\end{enumerate}

The deep ensemble size is 20. Hyper-parameters of each neural network in the ensemble (hidden layer size, number of hidden layers and alpha-dropout as special version of dropout in SNN \cite{Klambauer2017}) are the same. They are selected using a hybrid approach: first automatically, and then manually. The automatic approach is based on a Bayesian optimization with Gaussian Process regression \cite{Frazier2018ATO}, and it discovered the following hyper-parameter values for training the neural network in a single-task mode:

\begin{itemize}
    \item hidden layer size is 512;
    \item number of hidden layers is 12;
    \item alpha-dropout is 0.0003.
\end{itemize}

\begin{figure*}[hbt!]
 \centering
\begin{minipage}[h]{0.9\linewidth}
  \centering
  \centerline{\includegraphics[width=9cm]{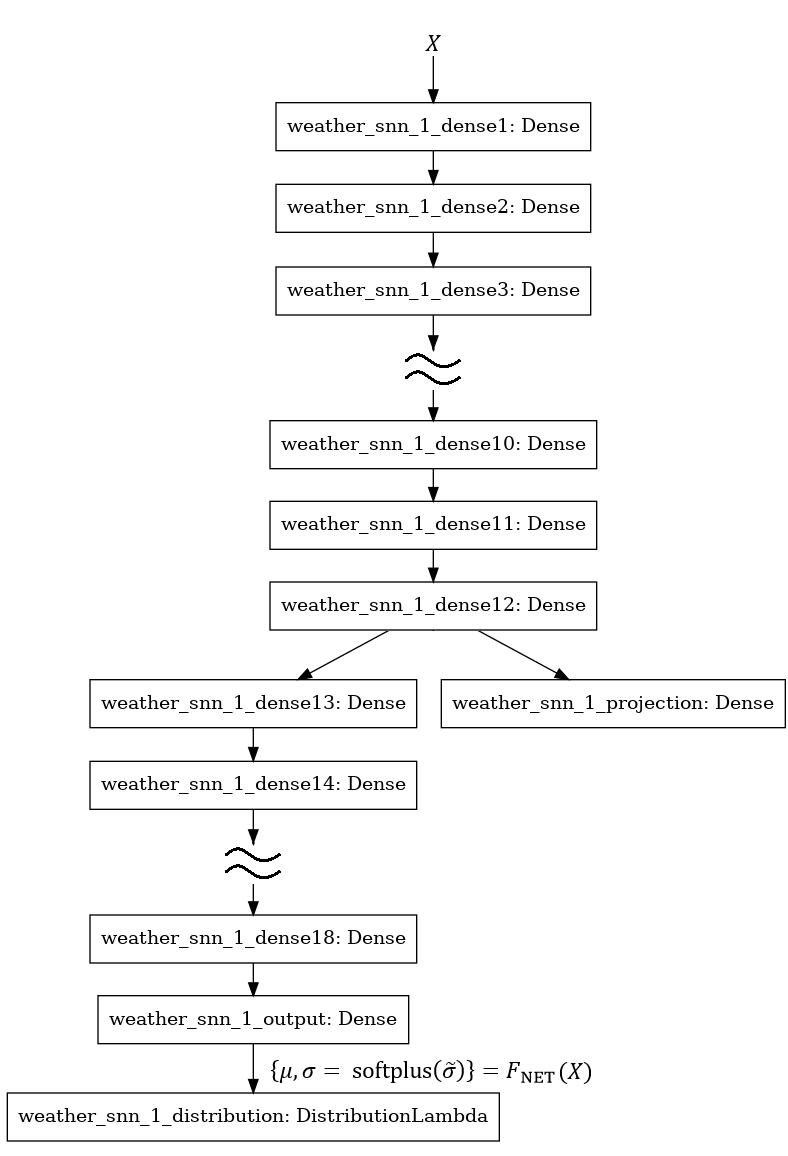}}
  \vspace{-0.1cm}
\end{minipage}
 \caption{Architecture of the deep neural network with hierarchical multitask learning.}
 \label{fig:nn-architecture}
\end{figure*}

After the implementation of the hierarchical multitask learning, the depth is manually increased up to 18 hidden layers: the low-level task (classification or supervised constrastive learning) is added after the 12th layer, and the high-level task (regression with uncertainty) is added after the 18th layer. General architecture of single neural network is shown on Fig.~\ref{fig:nn-architecture}. All "dense" components are feed-forward layers with self-normalization. Alpha-dropout is not shown to oversimplify the figure. The \texttt{weather\_snn\_1\_output} layer estimates the mean and the standard deviation of normal distribution, which is implemented using the \texttt{weather\_snn\_1\_distribution} layer. Another output layer named as \texttt{weather\_snn\_1\_projection} calculates low-dimensional projections for the supervised contrastive learning.

Rectified ADAM \cite{Liu2020OnTV} with Lookahead \cite{Zhang2019LookaheadOK} is used for training with the following parameters: learning rate is 0.0003, synchronization period is 6, and slow weights step size is 0.5. You can see the more detailed description of other training hyper-parameters in the Jupyter notebook \texttt{deep\_ensemble\_with\_uncertainty\_and\_spec\_fe.ipynb}. \footnote{This notebook is available at \url{https://github.com/bond005/yandex-shifts-weather}}

\subsection{Experiments}

Experiments were conducted according to data partitioning in section \ref{baseline}. Development and evaluation sets were not used for training and for hyper-parameter search. The quality of each modification of the method was first estimated on the development set, because all participants had access to the development set targets. After the preliminary estimation, the selected modification of the method was submitted to estimate R-AUC MSE on the evaluation set with targets concealed from participants.

In comparison with the baseline, the quality of the deep learning method is better (i.e. R-AUC MSE is less) on both datasets for testing:

\begin{itemize}
    \item the development set (for preliminary testing):
        \begin{enumerate}
            \item \textbf{proposed deep ensemble = 1.015};
            \item baseline (CatBoost ensemble) = 1.227;
        \end{enumerate}
    \item the evaluation set (for final testing):
        \begin{enumerate}
            \item \textbf{proposed deep ensemble = 1.141};
            \item baseline (CatBoost ensemble) = 1.335.
        \end{enumerate}
\end{itemize}

\begin{figure*}[hbt!]
 \centering
\begin{minipage}[h]{0.8\linewidth}
  \centering
  \centerline{\includegraphics[width=6.5cm]{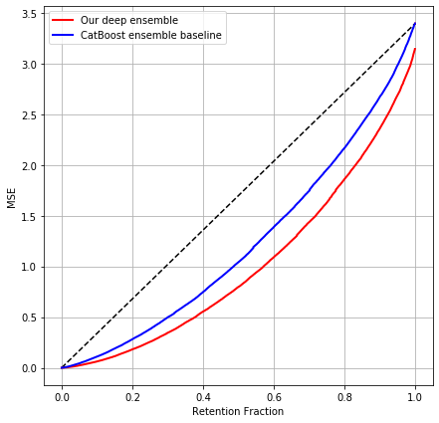}}
  \vspace{-0.1cm}
\end{minipage}
 \caption{Error retention curves on the development set.}
 \label{fig:devset-results}
\end{figure*}

Also, the results of the deep learning method are better with all possible values of the uncertainty threshold for retention (see Fig.~\ref{fig:devset-results}).

The total number of all submitted methods at the evaluation phase is 73. Six selected results (top-5 results of all participants and the baseline result) are presented in Table.~~\ref{tab:evalset-results}. The first three places are occupied by the following modifications of the proposed deep learning method:

\begin{itemize}
    \item \verb+SNN Ens U MT Np SpecFE+ is the final solution with "all-inclusive";
    \item \verb+SNN Ens U MT Np v2+ excludes the feature quantization;
    \item \verb+SNN Ens U MT+ excludes the feature quantization too, and classification is used instead of supervised contrastive learning as the low-level task in the hierarchical multitask learning.
\end{itemize}

\begin{table}[h!]
  \caption{Selected final results on the evaluation set. The author's team is \textbf{bond005}, and the baseline is developed by the \textbf{Shifts Team}}
  \label{tab:evalset-results}
  \centering
  \begin{tabular}{llll}
    \toprule
    Rank & Team    & Method                        & R-AUC MSE     \\
    \midrule
    1    & bond005 & \verb+SNN Ens U MT Np SpecFE+ & 1.1406288012  \\
    2    & bond005 & \verb+SNN Ens U MT Np v2+     & 1.1415403291  \\
    3    & bond005 & \verb+SNN Ens U MT+           & 1.1533415892  \\
    4    & CabbeanWeather & \verb+Steel box v2+    & 1.1575201873  \\
    5    & KDDI Research & \verb+more seed ens+    & 1.1593224114  \\
    55   & Shifts Team & \verb+Shifts Challenge+   & 1.3353865316  \\
    \bottomrule
  \end{tabular}
\end{table}

\section{Discussion}
\label{discussion}

Why is the neural network method better? Is its advantage randomness (particularly, task specifics), or does it have some fundamental basis? And lastly, what is the rationale for certain techniques in this method? These are very interesting questions, and they need to be discussed, of course.

\subsection{Why deep learning? Problems of decision trees and their ensembles}

Decision tree, including gradient boosted decision tree, is a very popular method to build classifiers and regressors, if the data is structured, but does not have local ordering. But any gradient boosted decision tree, like any other algorithm with decision trees, has two significant problems:

\begin{enumerate}
    \item The decision tree is built by a greedy algorithm.
    \item The decision tree cannot extrapolate effectively.
\end{enumerate}

The first problem is as follows. Base decision trees \cite{Quinlan1986} and their boosted ensembles \cite{friedman2001} use a \textbf{greedy algorithm} for recursive partitioning of the source dataset and all sub-sets, generated step-by-step. Any greedy algorithm is less effective in finding the optimal solution than gradient optimization, including gradient-based methods of deep learning.

\begin{figure*}[hbt!]
 \centering
\begin{minipage}[h]{0.97\linewidth}
  \centering
  \begin{subfigure}[b]{0.321\textwidth}
    \centering
    \includegraphics[width=4cm]{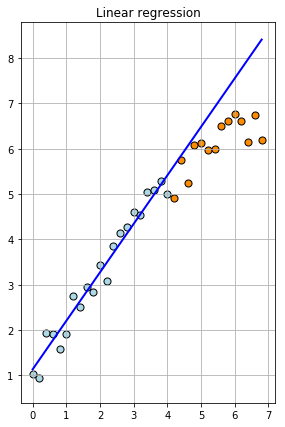}
    \caption{}
    \label{fig:regression-behaviour}
  \end{subfigure}
  \begin{subfigure}[b]{0.321\textwidth}
    \centering
    \includegraphics[width=4cm]{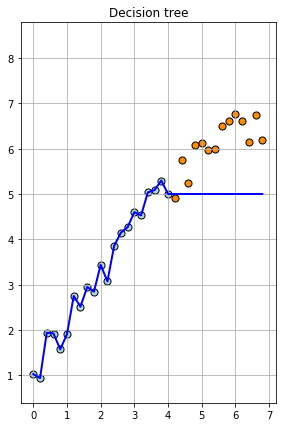}
    \caption{}
    \label{fig:tree-behaviour}
  \end{subfigure}
  \begin{subfigure}[b]{0.321\textwidth}
    \centering
    \includegraphics[width=4cm]{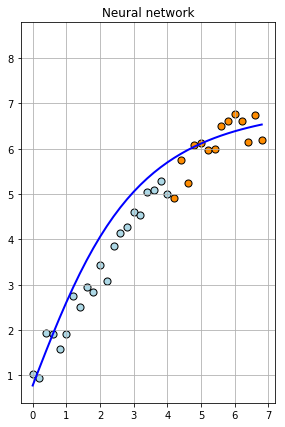}
    \caption{}
    \label{fig:dnn-behaviour}
  \end{subfigure}
  \vspace{-0.1cm}
\end{minipage}
 \caption{An example on synthetic data. Blue points are data for training and initial testing. Orange points are new data to predict. Shown: (a) predictions of linear regression; (b) predictions of decision tree; (c) predictions of neural network.}
 \label{fig:algorithm-behaviour}
\end{figure*}

At the heart of the second problem is common inefficiency of decision trees and several other "classic" ML algorithms (for example, nearest neighbors) in data extrapolation. In comparison with linear regression, decision tree can model nonlinear dependencies, but decision tree cannot do it for “out-of-distribution” data samples (illustration of this fact is shown on Fig.~\ref{fig:regression-behaviour} and \ref{fig:tree-behaviour}). As you can see, decision tree has better prediction quality in training and initial testing (blue points) than linear regression. But this powerful nonlinear algorithm loses catastrophically compared to linear regression, if the input data is new, i.e. out-of-distribution.

\subsection{The necessity of neural networks and more layers}

Applying deep neural network to the tabular data processing is conditional on following reasons.

Firstly, a neural network is an end-to-end pipeline of hierarchical feature extractors and a final estimator, and all pipeline components (i.e. layers) are trained together to minimize a task-conditioned loss function by gradient-based methods.

Secondly, a neural network is an effective form of non-linear regression, which has less problems with extrapolation and “out-of-distribution” data samples. Comparison of linear regression, decision tree and neural network with 3 layers is illustrated on the basis of toy regression example (see Fig.~\ref{fig:algorithm-behaviour}). Both decision tree (\ref{fig:tree-behaviour}) and neural network (\ref{fig:dnn-behaviour}) have better prediction quality in training and initial testing than linear regression (\ref{fig:regression-behaviour}). But neural network retains this quality on "out-of-distribution" data too.

However, until recently, fully-connected feed-forward neural networks were less used for tabular data processing in comparison with decision trees, than convolutional neural networks for computer vision and recurrent neural network for sequental data. And neural networks have some problems, first of all:

\begin{itemize}
    \item vanising gradient problem;
    \item bias-variance tradeoff;
    \item uncertainty and robustness.
\end{itemize}

\subsection{Vanishing gradient problem}
\label{vanishing}

The chain rule used in gradient calculation by the backprop can lead to very small values of gradient and thereby hinder weight updating. This problem is known as vanishing gradient. Normalization of all signals in the neural networks helps to solve this problem. This technique was analyzed in \cite{lecun98backprop}, and as a result the following techniques were proposed:

\begin{enumerate}
    \item All feature vectors should be standardized and decorrelated (removal of linear correlations can be implemented with data projecting to principal components).
    \item If solved task is regression, then all targets should be standardized too.
    \item Activation functions of hidden neurons should have a bipolar nonlinearity (for example, hyperbolic tangent or another similar function) instead of unipolar one (logistic sigmoid).
\end{enumerate}

But the third technique is not effective in a very deep neural network, because sigmoid-like activation function has a small derivative for non-small absolute value of argument, and it causes the vanishing gradient problem again. Applying piecewise linear function, like rectified linear unit \cite{glorot2011relu}, partially solves the small derivative problem (at least, for positive arguments), however, it generates yet another problem, connected with the systematic shift of hidden signals. There are several techniques of the "forced" normalization that were proposed to solve this problem, such as BatchNorm \cite{ioffe2015batchnorm}, LayerNorm \cite{Ba2016LayerN} and WeightNorm \cite{Salimans2016WeightNA}. BatchNorm (batch normalization) is popular in convolutional neural networks (CNNs), and LayerNorm (layer normalization) and WeightNorm (weight normalization) are suitable techniques for recurrent neural networks (RNNs) too. But training with abovementioned techniques of “forced” normalization is perturbed by all kinds of training stochasticity: stochastic gradient descent (SGD), stochastic regularization (like dropout), and the estimation of the normalization parameters. This problem is not critical for CNNs and RNNs, because their weights are shared (by receptive fields or by time). In contrast, fully-connected feed-forward neural networks (FFNNs) trained with “forced” normalization techniques suffer from these perturbations and have high variance in the training error. So, the special construction of self-normalizing neural networks, proposed in \cite{Klambauer2017}, is more efficient for FFNNs, including the case of tabular data processing.

Self-normalization is based on three principles:

\begin{enumerate}
    \item \textbf{Scaled exponentional linear unit} (SELU) as activation function, which implements the following nonlinearity after summator output \(s\):

\begin{displaymath}
f(s) = 1.0507 \cdot
  \begin{cases}
    s                           & \quad \text{if } s >    0 \\
    1.6733 \cdot e^{s} - 1.6733 & \quad \text{if } s \leq 0
  \end{cases}
\end{displaymath}

    \item Special initialization known as \textbf{Lecun initialization}: initial weights of neurons are sampled from the normal distribution \(\mathcal{N}(0, \sqrt{\frac{1}{n_{in}}})\), where \(n_{in}\) is the number of the neuron inputs.
    \item New dropout technique named as \textbf{“alpha dropout”}, which is based on randomly setting activations to the negative saturation value using the multiplicative noise.
\end{enumerate}

A detailed mathematical proof of the self-normalization properties in such neural network is presented in the appendix to the paper \cite{Klambauer2017}. Also, high effectiveness of self-normalizing networks in comparison with residual connections and different forms of the "forced" normalization is achieved on the experimental data from UCI \cite{dua2019}.

\subsection{Bias–variance tradeoff in deep learning}

The classic bias-variance problem is that increasing the ML algorithm complexity decreases the bias and increases the variance (this ML algorithm starts to approach overfitting) \cite{geman1992}. Therefore, any researcher who creates a deep learning method has to find a compromise solution, i.e. the happy medium between underfitting and overfitting, in some way. And the depth was usually reached at the expense of fully connected layers disavowal to limit the neural network's complexity. For example, it was implemented on the basis of restriction of inter-layer connections using receptive fields and shared weights in the convolutional neural networks.

In recent years, research of the bias-variance tradeoff concerning deep learning has received a new development. According to \cite{Neal2019}, increasing the deep learning algorithm complexity (in terms of the hidden layer size) decreases both the bias and the variance. At that, there are two kinds of variance:

\begin{itemize}
    \item \textit{variance due to optimization}: different initial weights lead to different training results;
    \item \textit{variance due to sampling}: different data subsets lead to different training results.
\end{itemize}

Joint variance is the combination of these two kinds of variance. And this joint variance has a unimodal dependence on the neural network complexity formulated as the hidden layer size. And interestingly, the variance due to sampling demonstrates a "traditional" behaviour, i.e. it slowly increases and eventually plateaus with complexity increasing, but the variance due to optimization at first increases and then decreases significantly. As a result, this determines the unimodality of the joint variance. Consequently, two conclusions follow on basis of the bias-variance tradeoff in deep learning:

\begin{enumerate}
    \item \textbf{Deep learning is a good} thing for any task, and increasing depth helps to partially solve the overfitting problem.
    \item Depth is not a silver bullet. It determines a lot, but not everything for overcoming the bias-variance tradeoff. Another part of solution can consist of more traditional techniques, such as \textbf{regularization and ensembles}.
\end{enumerate}

\subsection{Hierarchical multitask learning: why multitask and why hierarchical?}

Regularization is a well-known technique to increase the robustness of the trainable algorithm. Penalty is the simplest kind of regularization, and it is very efficient for linear models (for example, see \cite{bishop2006}). But deep neural network builds a hierarchical feature representation due its architecture and training, and regularization in deep learning should help to find a "good" hierarchy. Penalty is ineffective for solving such a problem.

There is yet another approach that can help to regularize the deep neural network. It is multi-task learning (MTL), and it is one of the most efficient techniques of "forcing" the network to search for an appropriate hierarchical system of the feature transformation. Multi-task learning introduces an inductive bias, based on a-priori relations between tasks: the trainable model is compelled to model more general dependencies by using the abovementioned relation as an important data feature \cite{caruana1997}.

Hierarchical MTL, in which different tasks use different levels of the deep neural network, provides more effective inductive bias compared to “flat” MTL. Also, hierarchical MTL helps to solve vanishing gradient problem in addition to the self-normalization, considered in subsection \ref{vanishing}. The idea of hierarchical MTL was first presented in \cite{Sgaard2016DeepML} as applied to natural language processing. Indeed, syntax and morphology are different levels of the natural language system, and syntax is based on morphology. Thus, such "natural" hierarchy can be projected onto a deep neural network trainable for natural language understanding.

Similarly to computational linguistics, the Weather Prediction task of the Yandex Shifts challenge can be decomposed into a hierarchical system of subtasks. Temperature prediction in the form of regression is a high-level fine-grained subtask with respect to the recognition of coarse temperature classes. Therefore, these two subtasks also form a "natural" hierarchy, which can be modeled using hierarchical MTL.

\subsection{Neural network: uncertainty and robustness}

Uncertainty modeling is one of the problems of neural networks. In the classification scenario, the neural network-based classifier is too “optimistic”, and it always returns a more “contrastive” probability distribution of classes. In the regression scenario, the regression model yields point predictions, and its training criteria (such as MSE or MAE) do not capture predictive uncertainty. Regression is a more interesting task in the context of the Weather Prediction on the Yandex Shifts challenge, and the uncertainty in the classification will not be analyzed further.

There are two key approaches to account and predict uncertainty with neural networks:

\begin{enumerate}
    \item Bayesian approach, which models epistemic uncertainty using a deep ensemble or its approximation, such as a Bayesian neural network. As is well known, this network is an approximation of an infinite ensemble \cite{Blundell2015}. The epistemic uncertainty is "model-driven", and it is related to the model behaviour (particularly, with its generalization ability).
    \item non-Bayesian approach, which models aleatoric uncertainty by means of direct estimation of normal distribution parameters. The aleatoric uncertainty is "data-driven", i.e. it is caused by incorrigible errors in measuring the target value. But these errors are distributed normally, and it is possible to reconstruct parameters of this distribution with maximum likelihood \cite{Nix1994EstimatingTM}.
\end{enumerate}

In practice, joint uncertainty includes both types of uncertainty (epistemic and aleatoric). So, accounting for their combination using the ensemble of deep regressors with the special loss function similar to described in the formula (\ref{eq:loss}) makes sense. The deep ensemble is a form of the Bayesian approach, and the loss function as a negative log-likelihood of parameters estimation of the Gaussian target error distribution embodies the non-Bayesian approach. An analogous technique is proposed in \cite{uncertainty2017}, but the authors contrapose the non-Bayesian approach to the Bayesian one, interpreting the Bayesian approach as a Bayesian neural network only. However, the ensemble of usual (non-Bayesian) neural networks is related to the Bayesian neural network: this is a technique of deep ensemble approximation \cite{Blundell2015, Wilson2020BayesianDL}.

\section{Conclusion}

There is every reason to believe that deep learning is the most efficient framework for the tabular data processing, and "classical" ML will become inferior to DL-based approaches in this domain, as it happened earlier in computer vision, for example. More layers provide more quality. Some problems connected with depth are solved using self-normalization and hierarchical multitask learning. Uncertainty can be modeled with combination of Bayesian and non-Bayesian techniques. Several variants of the algorithm built on the basis of these affirmations became winners of the Weather Prediction track of the Yandex Shifts challenge. This result could be considered as the confirmation of the effectiveness of deep learning, but so far this conclusion seems too optimistic. Really, this result inspires confidence in the correctness of the "deep way" and "more layers". But for the final confirmation, it is necessary to conduct a number of comparative experiments with various datasets devoted to classification and regression, including datasets with distributional shift for more robustness and more effective generalization evaluation. Besides, this paper does not consider the explaination problem of the deep learning algorithm, and an algorithm explainability can be very important in the real world cases. Therefore, these aspects should be used as directions for future research.

\begin{ack}
The author thanks Dr. Evgenii Vityaev for inspirational discussions about relations between deep learning, probability theory and formal logic.

Especially the author would like to thank his wife Viktoria Kondrashuk for her love, patience and constant support.

\end{ack}

\bibliographystyle{IEEEtranN}
\bibliography{references}

\end{document}